\documentclass[conference, letterpaper]{IEEEtran}
\ifCLASSINFOpdf
\else
\fi
\ifCLASSINFOpdf
   \usepackage[pdftex]{graphicx}
\else
\fi

\usepackage[cmex10]{amsmath}
\usepackage{color}
\usepackage{fancyhdr}
\usepackage[caption=false,font=footnotesize]{subfig}
\usepackage{rotating}
\usepackage{multirow}
\usepackage{comment}
\usepackage{booktabs}
\renewcommand{\thispagestyle}[2]{}

\fancypagestyle{plain}{
        \fancyhead{}
        \fancyhead[C]{first page center header}
        \fancyfoot{}
        \fancyfoot[C]{first page center footer}
}
\begin{document}

\title{Investigating Retrieval-Augmented Generation in Quranic Studies: A Study of 13 Open-Source \\Large Language Models}

\author{\IEEEauthorblockN{Zahra Khalila\IEEEauthorrefmark{1},
Arbi Haza Nasution\IEEEauthorrefmark{1}, Winda Monika\IEEEauthorrefmark{4}, Aytug Onan\IEEEauthorrefmark{5}, Yohei Murakami\IEEEauthorrefmark{6}, \\Yasir Bin Ismail Radi\IEEEauthorrefmark{7}, Noor Mohammad Osmani\IEEEauthorrefmark{10}}% and
%Nazlia Omar\IEEEauthorrefmark{9}}
\IEEEauthorblockA{\IEEEauthorrefmark{1}Department of Informatics Engineering, Universitas Islam Riau, Pekanbaru 28284, Indonesia\\
\IEEEauthorrefmark{4}Department of Library Information, Universitas Lancang Kuning, Riau 28266, Indonesia\\
\IEEEauthorrefmark{5}Department of Computer Engineering, College of Engineering and Architecture, \\Izmir Katip Celebi University, Izmir, 35620 Turkey\\
\IEEEauthorrefmark{6}Faculty of Information Science and Engineering, Ritsumeikan University,\\ Kusatsu, Shiga 525-8577, Japan\\
\IEEEauthorrefmark{7}Faculty of Al-Quran \& Sunnah, Universiti Islam Antarabangsa Tuanku Syed Sirajuddin (UniSIRAJ), \\Kuala Perlis, Perlis 02000, Malaysia\\
\IEEEauthorrefmark{10}Department Of Qur'an And Sunnah Studies, Ahas Kirkhs, \\International Islamic University Malaysia, Malaysia\\
%\IEEEauthorrefmark{9}Faculty of Information Science and Technology, Centre for Artificial Intelligence Technology,\\ Universiti Kebangsaan Malaysia, Bangi 43600, Malaysia\\
Email: arbi@eng.uir.ac.id}

}


%Telephone: (800) 555--1212, Fax: (888) 555--1212}


\maketitle

\begin{abstract}
%\boldmath
Accurate and contextually faithful responses are critical when applying large language models (LLMs) to sensitive and domain-specific tasks, such as answering queries related to quranic studies. General-purpose LLMs often struggle with hallucinations, where generated responses deviate from authoritative sources, raising concerns about their reliability in religious contexts. This challenge highlights the need for systems that can integrate domain-specific knowledge while maintaining response accuracy, relevance, and faithfulness. In this study, we investigate 13 open-source LLMs categorized into large (e.g., Llama3:70b, Gemma2:27b, QwQ:32b), medium (e.g., Gemma2:9b, Llama3:8b), and small (e.g., Llama3.2:3b, Phi3:3.8b). A Retrieval-Augmented Generation (RAG) is used to make up for the problems that come with using separate models. This research utilizes a descriptive dataset of Quranic surahs including the meanings, historical context, and qualities of the 114 surahs, allowing the model to gather relevant knowledge before responding. The models are evaluated using three key metrics set by human evaluators: context relevance, answer faithfulness, and answer relevance. The findings reveal that large models consistently outperform smaller models in capturing query semantics and producing accurate, contextually grounded responses. The Llama3.2:3b model, even though it is considered small, does very well on faithfulness (4.619) and relevance (4.857), showing the promise of smaller architectures that have been well optimized. This article examines the trade-offs between model size, computational efficiency, and response quality while using LLMs in domain-specific applications.
\end{abstract}



% \fi
\IEEEpeerreviewmaketitle

\section{Introduction}
Natural language processing (NLP) has been transformed as a result of the development of large language models (LLMs), which made it possible for these models to handle a wide range of activities. These include summarization and translation, as well as answering domain-specific questions \cite{raiaan2024review}. Further, these models can even serve as a good annotator for a number of NLP tasks \cite{nasution2024chatgpt}. Recent studies have explored the use of NLP in various domain-specific including Quranic studies, legal system, and medical field focusing on developing question-answering system. Alnefie et al. (2023) \cite{alnefaie2023gpt} has evaluated the effectiveness of GPT-4 in answering Quran-related questions and highlighting challenges in context understanding and answer accuracy. However, their work is limited by its reliance on a general-purpose LLM without domain-specific fine-tuning, which affects responce precision for nuance religious queries.  
Retrieval-Augmented Generation (RAG) has been succesfully applied in general knowledge such as medical domains to mitigate these issues \cite{zhou2024gastrobot}, and legal domain such as research conducted by Pipitone and Alami (2024) \cite{pipitone2024legalbench} introduced LegalBench-RAG to evaluate retrieval accuracy in legal question-answering tasks.
While significant progress has been made across these domains, challenges related to data quality, retrieval precision, domain-specific adaptation, and computational efficiency persist. This study builds upon these works by benchmarking open-source LLMs using the RAG framework to address the challenges in Quranic knowledge retrieval \cite{alrayzah2023challenges} while highlighting the balance between model size, performance, and efficiency. Using LLMs in religious or culturally sensitive environments has certain challenges \cite{fan2024survey}, including ensuring the accuracy, contextual relevance, and authenticity of the generated responses. These issues are particularly significant when engaging with content derived from religious texts, as distortion or hallucination may result in misunderstandings and a loss of faith in AI systems \cite{sun2024redeep}.

This paper examines the role of LLMs in quranic studies \cite{patel2023building}. We use a dataset from a book about the 114 surahs of the Qur'an \cite{AlBitaqat2023}, rather than the Qur'an text itself. This dataset provides thorough information about each surah, including meaning, provenance of revelation, and historical context. Descriptive insights are essential for offering relevant and accurate answers to questions regarding quranic studies.

Since Qur'an is the sacred revelation that remains intact in today’s world without any human involvement, and the Qur’an is the only source to link human beings with their creator, an attempt to identify the reliable and trustworthy LLMs is really important. As they provide useful resources for the readers, accuracy and truthfulness must be given due importance while exploring the provided information from such sources. 

In order to overcome the difficulties associated with hallucination and accuracy, we have implemented a framework known as Retrieval-Augmented Generation (RAG) \cite{njeh2024enhancing,lewis2020retrieval,brown2020language}. This method integrates LLM semantics with a vector database to retrieve relevant data from descriptive datasets \cite{gao2023retrieval,kamalloo2023evaluating}. The RAG method guarantees that responses derive from authoritative sources that are contextually appropriate, thereby reducing the probability of delivering content that is unsubstantiated or irrelevant. Citations are supplied in each response, which enables users to trace the information back to the descriptive dataset. This further enhances the level of trust and transparency \cite{zhou2024trustworthiness}.

The objectives of this research are threefold: (1) to compare 13 open-source LLMs in terms of their ability to respond accurately and faithfully to questions about quranic studies \cite{shabaz2023islamic}, (2) to assess the effectiveness of the RAG approach in reducing hallucination and ensuring response relevance \cite{gupta2024comprehensive},  and (3) to provide insights into the use of descriptive datasets for religious education and AI-based knowledge systems \cite{liu2024datasets}. The evaluation criteria consist of context relevance, response faithfulness, and answer relevance, and they are evaluated using human evaluation. \cite{gao2023retrieval}.

This study provides a robust framework for integrating LLMs with descriptive datasets, thereby contributing to the expanding field of domain-specific AI applications \cite{rizqullah2023qasina}. It emphasizes the strengths and limitations of current LLMs in managing sensitive religious topics and establishes a foundation for future developments in AI-driven educational and informational tools.

The rest of this paper is organized as follows: Section 1 introduces the challenges of using LLMs for Quranic studies and outlines the research objectives. Section 2 reviews related work, discussing previous studies on LLM applications in religious text analysis and RAG-based retrieval systems. Section 3 provides a detailed description of the experimental setup, covering the dataset, NLP tasks and evaluation guidelines, dataset selection and curation, human evaluators, metrics for quality evaluation, large language models, and hardware and software configuration. Section 4 presents the experimental results of various LLM models. Section 5 discusses key findings, including performance insights based on model size, the effectiveness of the RAG framework, the trade-off between computational resources and response quality, the surprising performance of Llama3.2:3b, and implications for domain-specific tasks. Section 6 concludes the paper by summarizing the main contributions and providing suggestions for future research.

\section{Materials and Methods}
This section outlines the methodical strategy employed in our research. We begin by analyzing the dataset, detailing its source, structure, and descriptive content. Subsequently, the system's responsibilities are thoroughly delineated, encompassing the formulation of solutions to user concerns concerning Islamic doctrines \cite{abubakari2024evaluating}. In addition, we provide a summary of the rules that have been set for human evaluators who are responsible for evaluating the outputs of the system. To ensure that the evaluations are reliable and consistent, we have produced these guidelines.

\subsection{Datasets}
The dataset used in this research comes from a descriptive book providing a thorough study of the 114 surahs (Chapters) of the Qur'an. In this preliminary study, 20 surahs were selected and tested out of the total 114 surahs. The dataset includes numerous descriptive elements for each surah, such as but not limited to:

\begin{itemize}
    \item \textbf{Number of Verses}: The total number of verses in each surah.
    \item \textbf{Meaning of its Name}: A description of the surah's title and its importance.
    \item \textbf{Reason for its Name}: An explanation for the designation of the surah, frequently linked to its subject matter or motifs.
    \item \textbf{Names}: Alternative titles or names linked to the surah, if relevant.
    \item \textbf{General Objective}: A brief explanation of the primary message or objective of the surah.
    \item \textbf{Reason for its Revelation}: The circumstances or context in which the surah was revealed, as available.
    \item \textbf{Virtues}: Key benefits or spiritual rewards associated with reciting or understanding the surah, often supported by hadith (sayings of the Prophet Muhammad, peace be upon him).
    \item \textbf{Relationships}: Insights into the connections between the beginning and end of the surah, or its relationship to preceding or succeeding chapters.
\end{itemize}
For instance, Surah Al-Hadid (Chapter 57) is described as follows:
\begin{itemize}
    \item \textbf{Number of Verses}: 29.
    \item \textbf{Meaning of its Name}: "Al-Hadid" translates to "The Iron" in Arabic.
    \item \textbf{Reason for its Name}: It is the only chapter where the benefits of iron are mentioned, symbolizing strength and utility.
    \item \textbf{General Objective}: Encourages the virtue of spending in the cause of Allah as an appreciation of His favors.
    \item \textbf{Virtues}: Includes a hadith where the Prophet Muhammad (peace be upon him) recommends reciting this chapter as one of three glorifications of Allah.
    \item \textbf{Relationships}: Highlights thematic continuity between its verses and connections to preceding chapters, such as Surah Al-Waqi'ah.
\end{itemize}
%\subsubsection{Dataset Structuring and Vectorization}

The dataset was preprocessed and organized into structured fields to ensure efficient retrieval and usability in the study. Key steps included:
\begin{itemize}
    \item \textbf{Field Segmentation}:
    Each descriptive element (e.g., "virtues," "relationships") was extracted and stored as a separate field for better semantic alignment with queries.
    \item \textbf{Vectorization}:
    Text data was transformed into high-dimensional vector embeddings through the use of cutting-edge NLP models, which facilitated the search for semantic similarity \cite{han2023comprehensive}.
    \item \textbf{Storage in a Vector Database}:
    The organized and scalar data was kept in a scaled vector database so that it would be easy to find the right descriptions during query processing \cite{jing2024large},\cite{han2023comprehensive}.
\end{itemize}
%\textbf{Significance of the Dataset}

This particular dataset provides lots of information that goes beyond the actual text of the Qur'an, including contextual and interpretative details. Based on this, it is a great instrument for evaluating the ability of LLMs to produce responses that are true, accurate, and relevant to the context in which they are being used. By emphasizing descriptive elements, the dataset ensures that responses align with established interpretations and scholarly perspectives.

\subsection{NLP Tasks and Evaluation Guidelines}
A Retrieval-Augmented Generation (RAG) architecture will be utilized in order to accomplish the objective of this study, which is to evaluate large language models (LLMs) in the context of answering questions related to quranic studies \cite{siriwardhana2023improving}. The RAG approach combines LLMs with semantic retrieval to provide contextually relevant and authoritative responses from a descriptive dataset. The primary tasks and evaluation guidelines used to assess the system's performance are outlined in full below \cite{gao2023retrieval}.

\subsubsection{NLP Tasks}
The research's primary NLP task is to generate semantically pertinent and contextually accurate responses to inquiries regarding quranic studies. The system employs a Retrieval-Augmented Generation (RAG) architecture, combining retrieval-based and generative methodologies, to ensure that responses are both dataset-based and linguistically coherent. The system executes the following tasks:
\begin{itemize}
    \item \textbf{Semantic Search and Retrieval:} Upon a user's query submission, the system does a semantic similarity search over the vectorized dataset obtained from Qur’anic surah descriptions \cite{liu2024datasets}. This procedure determines the most contextually pertinent entries from the dataset to respond to the query.
    \item \textbf{Response Generation:} The retrieved descriptions are submitted to the LLMs, which produce a comprehensive response \cite{gao2023retrieval}. This response integrates the retrieved information and provides explanatory content to address the query.
    \item \textbf{Citations and Contextualization:} Each generated response includes references to the original dataset entries (e.g., surah descriptions or specific virtues), allowing users to trace the information back to its source \cite{alan2024rag}.
\end{itemize}

\subsubsection{Evaluation Guidelines}
To assess the quality of the responses generated by the system, human evaluators followed a structured set of evaluation guidelines. These guidelines provided a consistent framework for scoring responses across three key dimensions: Context Relevance, Answer Faithfulness, and Answer Relevance \cite{gao2023retrieval}. Each dimension is explained below, along with its calculation method and examples.
\begin{itemize}
    \item \textbf{Context Relevance} evaluates how precisely the retrieved and generated responses align with the user query while avoiding irrelevant or extraneous information. The relevance score is calculated using the \textbf{precision@k} metric, where \( k \) represents the number of top retrieved results considered as shown in Equation~\ref{eq2}.
\begin{equation}
\label{eq2}
\small
\text{Precision@k} = \frac{\text{No. of relevant results in the top-k responses}}{k}    
\end{equation}

\textbf{Example:}  
\begin{itemize}
    \item \textbf{Query:} “What is the reason for Surah Al-Fatihah being named Umm Al-Kitab?”
    \item \textbf{Retrieved Information:} 
    \begin{enumerate}
    \item Surah Al-Fatihah is named Umm Al-Kitab because it summarizes the essence of the Qur'an (relevant).
    \item It is recited in every unit of prayer (relevant).
    \item Surah Al-Baqarah discusses laws and stories (irrelevant).
    \item Surah Al-Fatihah has seven verses (relevant).
    \item Surah An-Nas is the last chapter of the Qur'an (irrelevant).
\end{enumerate}
\end{itemize}

If \( k = 5 \), then 3 out of the 5 retrieved results are relevant:
\[
\text{Precision@5} = \frac{3}{5} = 0.6
\]

The context relevance score for this response is therefore 0.6.

    \item \textbf{Answer Faithfulness} ensures that the generated responses accurately represent the retrieved information without introducing unsupported content or hallucinations. Evaluators compare the generated response with the dataset to verify factual consistency.

    \textbf{Example:}
\begin{itemize}
    \item \textbf{Query:} “What does Surah Al-Fatihah emphasize?”
    \item \textbf{Retrieved Information:} Surah Al-Fatihah emphasizes monotheism, gratitude, and seeking guidance from Allah.
    \item \textbf{Faithful Response:} Surah Al-Fatihah highlights the themes of monotheism, gratitude, and the importance of seeking Allah's guidance.
    \item \textbf{Non-Faithful Response:} Surah Al-Fatihah emphasizes the stories of past prophets.
\end{itemize}

The faithful response adheres strictly to the retrieved information, while the non-faithful response introduces unsupported content.

    \item \textbf{Answer Relevance} measures whether the response directly addresses the query while maintaining semantic and theological appropriateness. It assesses completeness, clarity, and alignment with the question.
    
    \textbf{Example:}
\begin{itemize}
    \item \textbf{Query:} “Why is Surah Al-Fatihah called Umm Al-Kitab?”
    \item \textbf{Relevant Response:} Surah Al-Fatihah is called Umm Al-Kitab because it summarizes the central teachings of the Qur'an and is recited in every unit of prayer.
    \item \textbf{Irrelevant Response:} Surah Al-Fatihah has seven verses and is the first chapter of the Qur'an.
\end{itemize}

The relevant response directly answers the query, providing reasoning, while the irrelevant response, though factually correct, fails to address the specific question.

\end{itemize}

\subsubsection{Evaluation Process} {Human evaluators assessed the system-generated responses through a web-based platform, where they were presented with prompts, the corresponding responses, and an interface for scoring. The evaluation process included the following steps:}
\begin{itemize}
    \item\textbf{Reviewing Responses: } The process of reviewing responses was a critical step in evaluating the quality of the outputs generated by the large language models (LLMs). Human evaluators carried out this task through a web-based platform specifically designed to facilitate structured and unbiased assessments.
    \item\textbf{Scoring System:} For each response, evaluators assigned scores on a Likert scale (1 to 5) for the three evaluation criteria: Context Relevance, Answer Faithfulness, and Answer Relevance. In addition to numerical scores, evaluators could provide written feedback to justify their evaluations \cite{zhou2024trustworthiness}. This qualitative feedback emphasized specific faults or applauded features of the response, providing more insight into the system's performance.
    \item\textbf{Reevaluation and Calibration:} Since numerous replies were provided for each query, evaluators could compare the quality of outputs from various LLMs. This comparative approach was instrumental in identifying the models' relative strengths and limitations, thereby enabling a more thorough evaluation \cite{tam2022evaluating}. To verify the dependability of their judgments, evaluators returned to a subset of previously evaluated responses on a regular basis and reassessed them. This consistency check allowed evaluators to reflect on their scoring processes and ensure they were in line with the rating criteria.
\end{itemize}
   
The structured evaluation guidelines guaranteed that the assessment process was meticulous, consistent, and transparent. The guidelines established a comprehensive framework for assessing the system's performance by emphasizing context relevance, answer faithfulness, and answer relevance \cite{zhou2023context}. This method enable a thorough comparison of several LLMs and provided vital insights into their performance in answering Islamic queries with contextual precision and faithfulness \cite{kamalloo2023evaluating, zhou2023context}. The evaluations were submitted through the platform after all responses to a specific query were evaluated, scored, and commented on. In order to provide a comprehensive dataset for the purpose of investigating the LLMs, the platform logged and stored the data for research \cite{wang2024evaluating}.
   
\subsection{Dataset Selection and Curation}
{The dataset used in this study was carefully selected and organized to guarantee it aligns with the goals of assessing large language models (LLMs) within the framework of quranic studies. The process of selection and curation included identifying a reliable source, organizing the data, and confirming its alignment with the research goals} \cite{kamalloo2023evaluating, wang2024evaluating}.

\subsubsection{Selection Criteria}
The dataset was chosen according to these specific criteria:
\begin{itemize}
    \item \textbf{Authenticity:} { The source underwent a thorough review to confirm its compliance with recognized Islamic scholarship and the absence of speculative interpretations} \cite{tam2022evaluating, zhou2023context}.
    \item \textbf{Descriptive Richness:} {The dataset must deliver comprehensive, contextually rich descriptions that can be effectively employed for semantic search and response generation} \cite{zhou2023context, huang2024enhancing}.
    \item \textbf{Clarity and Accessibility:} {The content needed to be created in a structured and clear manner, facilitating both manual review and computational processing} \cite{kamalloo2023evaluating, wang2024evaluating}.
    \item \textbf{Relevance:} {The dataset was meticulously curated to facilitate the process of addressing inquiries related to quranic studies, emphasizing themes that are frequently observed in these discussions} \cite{tam2022evaluating}
\end{itemize}

\subsubsection{Curation Process}
A comprehensive curation procedure was carried out on the dataset in order to get it ready for integration with the retrieval-augmented generation (RAG) system and LLMs:
\begin{itemize}
    \item \textbf{Data Digitization:} The text from the source book was digitized to generate a dataset that can be accessed by machines. Optical Character Recognition (OCR) tools were employed where necessary to convert printed material into digital text \cite{kamalloo2023evaluating, wang2024evaluating}.
    \item \textbf{Data Structuring:} {The content was segmented into surah name, number of verses, reason for the name, general objective, virtues, and relationships. Each field was carefully labeled to facilitate precise retrieval} \cite{zhou2023context}.
    \item \textbf{Content Validation:} {The digitized and structured dataset was reviewed by experts in Islamic studies to verify its accuracy and alignment with the original source} \cite{tam2022evaluating}.
    \item \textbf{Preprocessing:} Unnecessary or redundant information was removed, and inconsistencies were corrected. Tokenization was performed to split the text into smaller, manageable units for processing by the semantic search system.
    \item \textbf{Vectorization:} {The structured data was transformed into high-dimensional vector embeddings using pre-trained language models \cite{monir2024vectorsearch}. This step allowed for efficient and accurate semantic similarity searches within the dataset}.
    \item \textbf{Storage in a Vector Database:} {The vectorized dataset was stored in a scalable and efficient vector database, enabling quick retrieval of relevant entries based on user queries} \cite{han2023comprehensive}.
\end{itemize}

\subsubsection{Dataset Integrity}
To ensure the integrity and reliability of the dataset, multiple layers of validation were employed, including manual review and automated consistency checks, regular audits of the data were conducted to identify and rectify any errors or discrepancies, and a backup of the raw and processed datasets was maintained for reproducibility and future reference.

\subsubsection{Strengths of the Dataset}
\begin{itemize}
    \item\textbf{Richness in Context:} {The dataset goes beyond literal translations, providing thematic, historical, and theological insights.}
    \item\textbf{High Relevance:} {The information directly supports answering user queries about quranic studies.}
    \item\textbf{Scalability:} {The vectorized format enables integration with modern NLP systems and future upgrades.}
\end{itemize}

This curated dataset ensures that the responses generated by the system are accurate, faithful, and contextually relevant, thereby serving as the foundation for the investigating and evaluation of the LLMs

\subsection{Human Evaluators}
In this research, human evaluators were instrumental in evaluating the responses produced by the large language models (LLMs) \cite{elangovan2024considers}. The evaluations were carried out using a specially designed website that focused on optimizing the evaluation process and maintaining consistency. The website offered evaluators with questions and responses from several LLMs, allowing for a systematic assessment using preset criteria which are context relevance, answer faithfulness, and answer relevance.

\subsubsection{Evaluator Selection} The evaluators were selected with careful consideration to guarantee that they had the proper knowledge and comprehension of quranic studies, given that the study centers on inquiries pertaining to Islamic content. Criteria for selection included: 
    \begin{itemize}
        \item {\textbf{Knowledge of quranic studies}: Evaluators who had either formal education or significant experience in Islamic studies were prioritized.}
        \item {\textbf{Analytical Skills}: In order to evaluate the quality of responses across multiple dimensions, evaluators were required to possess strong analytical skills.}
        \item {\textbf{Familiarity with Evaluation Tasks}: It was considered beneficial to have prior experience analyzing textual data or utilizing NLP technologies.}
    \end{itemize}
Evaluators from a variety of backgrounds were included to make sure the replies were evaluated fairly and without bias.

\subsubsection{Evaluation Platform}
The evaluation process was conducted through a dedicated website designed to facilitate efficient and user-friendly assessments. The platform included the following features: 

\begin{itemize}
    \item\textbf{Query-Response Display: }
        \begin{itemize}
            \item {Each evaluation session displayed a prompt (query) along with responses generated by different LLMs.}
            \item {Responses were anonymized to prevent bias, ensuring that evaluators were not influenced by the identity of the LLM responsible for generating a response.}
        \end{itemize}
    \item\textbf{Scoring Interface: }Evaluators rated each response based on the three evaluation criteria: 
        \begin{itemize}
            \item {Context Relevance: Precision and alignment of the response with the query.}
            \item {Answer Faithfulness: Accuracy of the response in relation to the retrieved dataset content.}
            \item {Answer Relevance: Appropriateness and direct pertinence of the response to the query.}
        \end{itemize}
    A Likert scale (1–5) was used for scoring, where 1 indicated poor performance and 5 indicated excellent performance.
     \item\textbf{Feedback Mechanism: }Evaluators could provide written comments to justify their scores or highlight specific issues in the responses. This feature allowed the identification of nuanced errors that might not be captured by numerical scores alone.
\end{itemize}

\subsubsection{Significance of Human Evaluations}
The use of human evaluators provided an essential layer of validation for the study, ensuring that the generated responses were assessed not only for technical accuracy but also for their theological and contextual integrity \cite{elangovan2024considers}. By leveraging a well-structured platform and robust evaluation criteria, the study ensured that the investigation of LLMs was both rigorous and comprehensive, offering valuable insights into their performance in responding to Islamic queries \cite{feng2024sample}.

\subsection{Metric for Quality Evaluation}
This study employs Inter-Evaluator Agreement (IEA) as the primary metric to ensure the quality, reliability, and consistency of human evaluations \cite{moons2023measuring}. Since human evaluators, rather than annotators, were tasked with assessing the generated responses, IEA provides a robust measure of agreement across evaluators, validating the credibility of the evaluation process and the results.

IEA measures the level of consistency between evaluators when scoring responses based on predefined criteria: context relevance, answer faithfulness, and answer relevance. An elevated IEA score signifies uniform application of evaluation criteria by assessors, whereas a diminished score reveals inconsistencies that may necessitate recalibration.  

Fleiss’ Kappa, as presented in Equation~\ref{eq1}, was employed to calculate IEA due to its appropriateness for evaluating agreement among multiple evaluators concurrently \cite{moons2023measuring}. The consistent monitoring of IEA scores allowed the research team to detect discrepancies promptly and facilitate recalibration sessions as needed, ensuring evaluators' comprehension and interpretation of the scoring guidelines were aligned. This methodical approach guaranteed that the evaluation process was dependable, replicable, and aligned with the study's goals.

\begin{equation}
\label{eq1}
\kappa = \frac{\bar{P} - \bar{P}_e}{1 - \bar{P}_e}
\end{equation}

\text{where:}
\begin{equation}
\bar{P} = \frac{1}{N} \sum_{i=1}^{N} P_i \quad \text{and} \quad \bar{P}_e = \sum_{k=1}^{K} p_k^2
\end{equation}

\text{with:}

\begin{equation}
P_i = \frac{1}{n(n-1)} \sum_{k=1}^{K} n_{ik} (n_{ik} - 1), \quad p_k = \frac{\sum_{i=1}^N n_{ik}}{Nn}
\end{equation}

\text{where:}
\begin{itemize}
    \item \( \kappa \): Fleiss' Kappa value.
    \item \( \bar{P} \): Average observed agreement across all items.
    \item \( \bar{P}_e \): Expected agreement based on chance.
    \item \( P_i \): Proportion of agreement for item \( i \).
    \item \( p_k \): Proportion of ratings in category \( k \).
    \item \( n \): Total number of ratings per item.
    \item \( n_{ik} \): Number of raters who assigned category \( k \) to item \( i \).
    \item \( N \): Total number of items.
    \item \( K \): Total number of categories.
\end{itemize}

\subsection{Large Language Models}
The efficacy of numerous large language models (LLMs) in responding to inquiries regarding quranic studies is assessed in this study using a Retrieval-Augmented Generation (RAG) framework \cite{islam2024open}. A comprehensive comparison of the capabilities of the LLMs selected for investigation is possible due to the fact that they represent a variety of architectures and parameter scales. More information about the models, how they are put together, and how they relate to this study is given below.

\subsubsection{Llama} 
Meta AI has developed the Llama (Large Language Model Meta AI) family of models, which are state-of-the-art transformer-based architectures that are optimized for natural language understanding and generation \cite{touvron2023llama}. Llama models are trained on vast, diversified corpora to do various NLP tasks such as contextual reasoning, question answering, and text summarization. They are available in a variety of parameter values, which provides a degree of flexibility in terms of computational requirements and performance. 
The Llama models were incorporated in this investigation due to their adaptability and superior performance across various parameter sizes. A thorough investigation of how model size affects the capacity to produce faithful, accurate, and contextually relevant replies was made possible by the range of configurations. The comparison of Llama generations (e.g., Llama3 with Llama3.1) yielded insights on the impact of incremental architectural enhancements on performance.

\subsubsection{Gemma} 
Google's DeepMind developed the Gemma family of large language models, which are a set of transformer-based designs that are best for understanding and creating natural language. The Gemma models are engineered to provide superior performance while ensuring efficiency, rendering them adaptable for various jobs. These models have undergone pre-training on varied and comprehensive datasets, enabling them to generalize effectively across multiple domains \cite{team2024gemma}. Gemma models are offered in many parameter scales, including 27b, 9b, and 2b, providing flexibility to optimize performance and computational demands. Their versatility renders them appropriate for applications from resource-intensive jobs to real-time implementations in limited surroundings.
\subsubsection{QwQ}
The QwQ model developed by Alibaba Cloud, which has 32 billion parameters, is a large-scale transformer-based language model designed to handle complex natural language processing tasks \cite{bai2023qwen}. While specific information about the QwQ model’s architecture or pre-training details is limited in comparison to more established models like Llama and Gemma, its parameter scale positions it as a powerful model capable of capturing complex relationships in textual data. Its large size enables it to perform effectively on tasks requiring nuanced comprehension, contextual reasoning, and content generation across a wide range of subjects. 
\begin{comment}
The QwQ:32b has a large amount parameters, allowing it for complicated query processing and contextual comprehension. Its size enables accurate, consistent, and thorough responses. QwQ models is great at capturing long-range dependencies and subtle patterns in input data, making them superior to smaller models. This makes it especially useful for activities that need precise reasoning, such as question answering, summarization, and information retrieval. The capacity to handle large-scale activities assures strong performance in cases with domain-specific or knowledge-intensive queries.

The QwQ:32b model has certain limits despite to its virtues. Its high parameter count generates major computing requirements that call for time for inference, memory, and processing capability. For systems with limited hardware resources or real-time applications, this model reduces efficiency. Furthermore, as with other huge models, especially when addressing unclear questions, QwQ could periodically provide responses that are verbose or include extraneous details. Retrieval-Augmented Generation (RAG) and other retrieval-based processes can help to reduce these problems and guarantee more concentrated results.
\end{comment}
\subsubsection{Phi} 
Microsoft created the Phi family of language models, which are a group of lightweight transformer-based models that work best for jobs that involve processing natural language. Even though the Phi models have lower parameter sizes compared to other large-scale models such as Llama and Gemma, they are engineered to exceed expectations, providing robust performance while ensuring computational efficiency \cite{abdin2024phi}. They are pretrained on rigorously selected datasets that prioritize high-quality information, enabling effective generalization across tasks despite a reduced number of parameters. This method guarantees that Phi models provide an exceptional equilibrium between performance and resource demands, rendering them especially appropriate for resource-limited settings and real-time applications.
\subsubsection{Key Features of the Models}
From smaller-scale models (e.g., 1 billion parameters) to large-scale ones (e.g., 70 billion parameters), the chosen models encompass a wide spectrum of parameter sizes. This variation enables a study of the trade-offs between response quality and computing efficiency. Smaller models might lose accuracy and contextual depth, yet producing faster responses with reduced processing expenses. On the other hand, it is anticipated that larger models will produce more nuanced and high-fidelity outputs, although at the expense of increased computational demands.

The models use transformer architectures, which are efficient in comprehending and producing natural language content. Their architecture allows to capturing the intricate patterns, correlations, and contextual subtleties within the dataset. This work emphasizes the models' capacity to adjust to specific domains, such as quranic studies, despite being trained on varied datasets. The assessment analyzes the efficacy of these models when enhanced with domain-specific data through the RAG methodology. The study evaluates the models in a zero-shot context, without any fine-tuning on the unique dataset. This method emphasizes the models' intrinsic capacity to generalize and appropriately respond by utilizing external information obtained from the descriptive dataset.

\subsubsection{Integration with the RAG Framework}
The selected LLMs were integrated with the RAG framework, allowing them to generate contextually relevant responses based on the dataset. The RAG framework enhances the performance of the models by providing contextual input, reducing hallucination, and facilitating citations.

\subsection{ Hardware and Software Configuration}
The experiments were conducted using a high-performance computing system equipped with an Intel(R) Xeon(R) Gold 5318Y CPU operating at 2.10 GHz with 24 cores. The system featured four NVIDIA RTX A6000 GPUs, each providing 48 GB of VRAM, enabling efficient handling of computationally intensive tasks, particularly those involving deep learning models. Additionally, the system was supported by 128 GB of RAM, ensuring smooth execution of memory-intensive operations and facilitating large-scale data processing. This configuration provided the computational resources necessary to run and evaluate the models effectively.
\section{Results}
The results of this study are based on the implementation of a RAG architecture, designed to evaluate the performance of 13 LLMs in answering questions related to Quranic studies. By leveraging a descriptive dataset of Quranic surahs, the RAG system facilitates the integration of external knowledge to address the limitations of standalone models. The comparative analysis focuses on assessing the relevance, faithfulness, and contextual accuracy of the responses generated by the LLMs within this framework.
\begin{figure}[!h]
  \centering
  \includegraphics[scale=0.8]{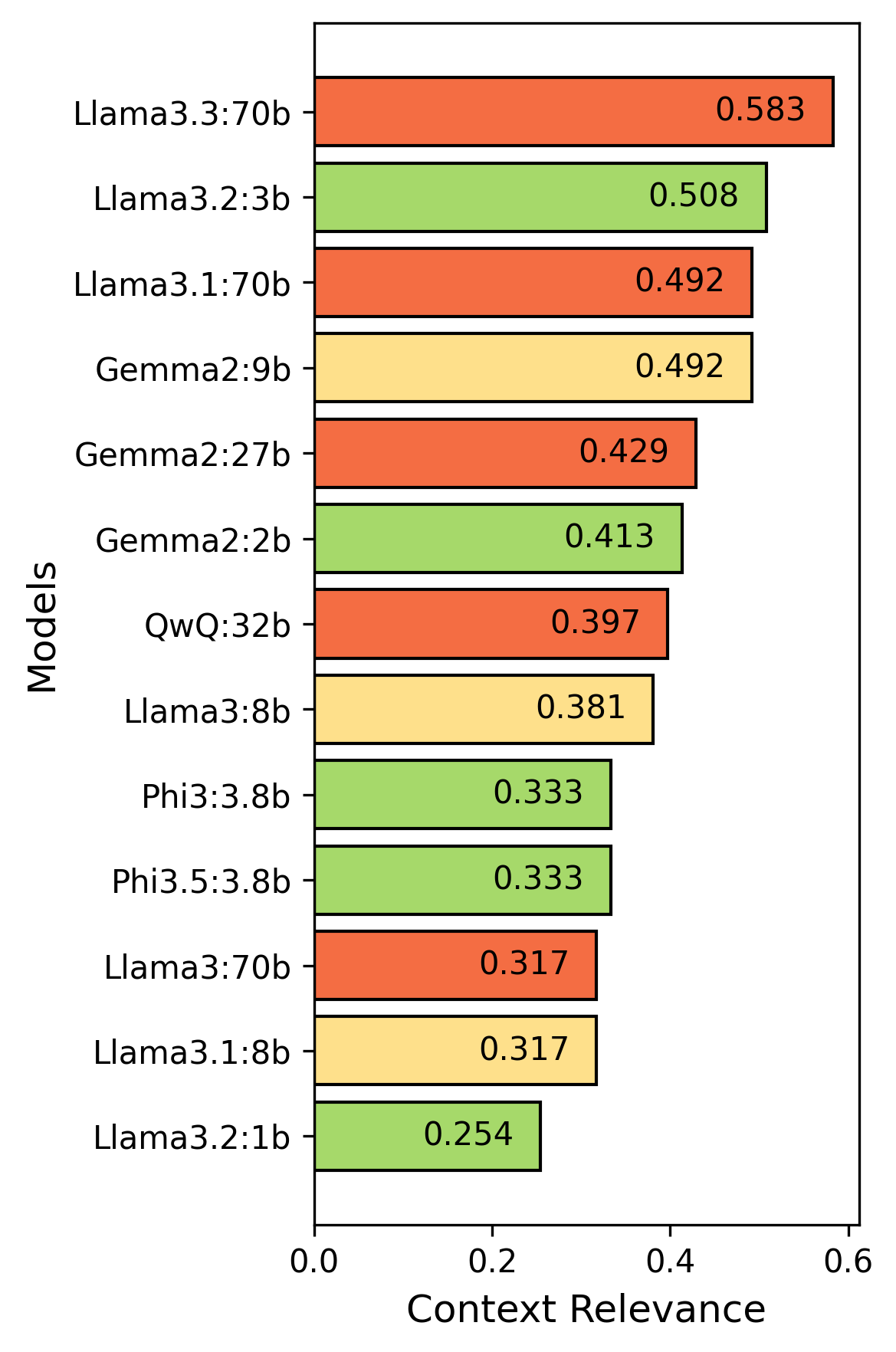} 
\caption{Context Relevance by the 13 LLMs.}
\label{fig.1}
\end{figure}

\begin{figure}[!h]
  \centering
  \includegraphics[scale=0.8]{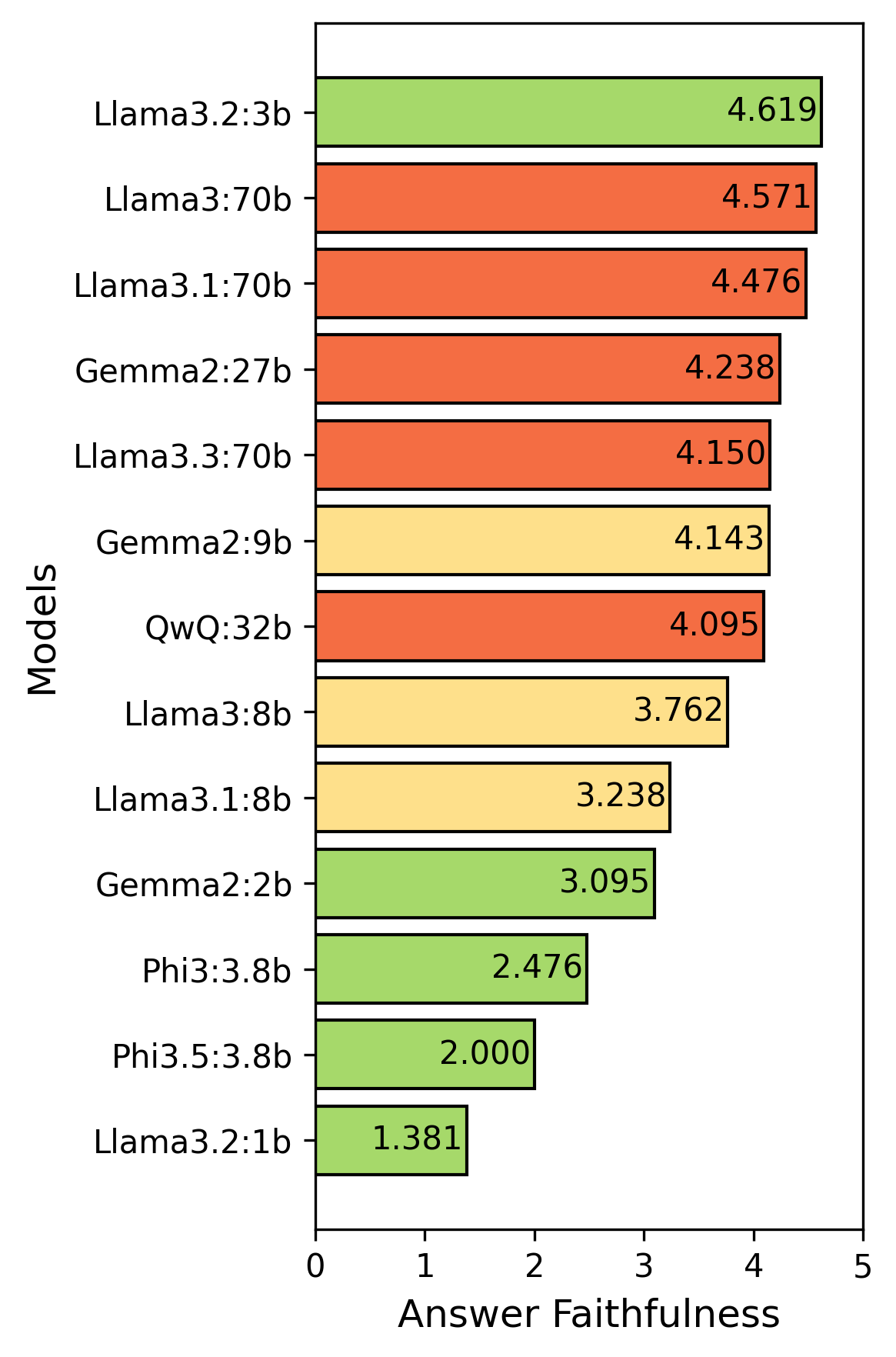} 
\caption{Answer Faithfulness by the 13 LLMs.}
\label{fig.2}
\end{figure}

\begin{figure}[!h]
  \centering
  \includegraphics[scale=0.8]{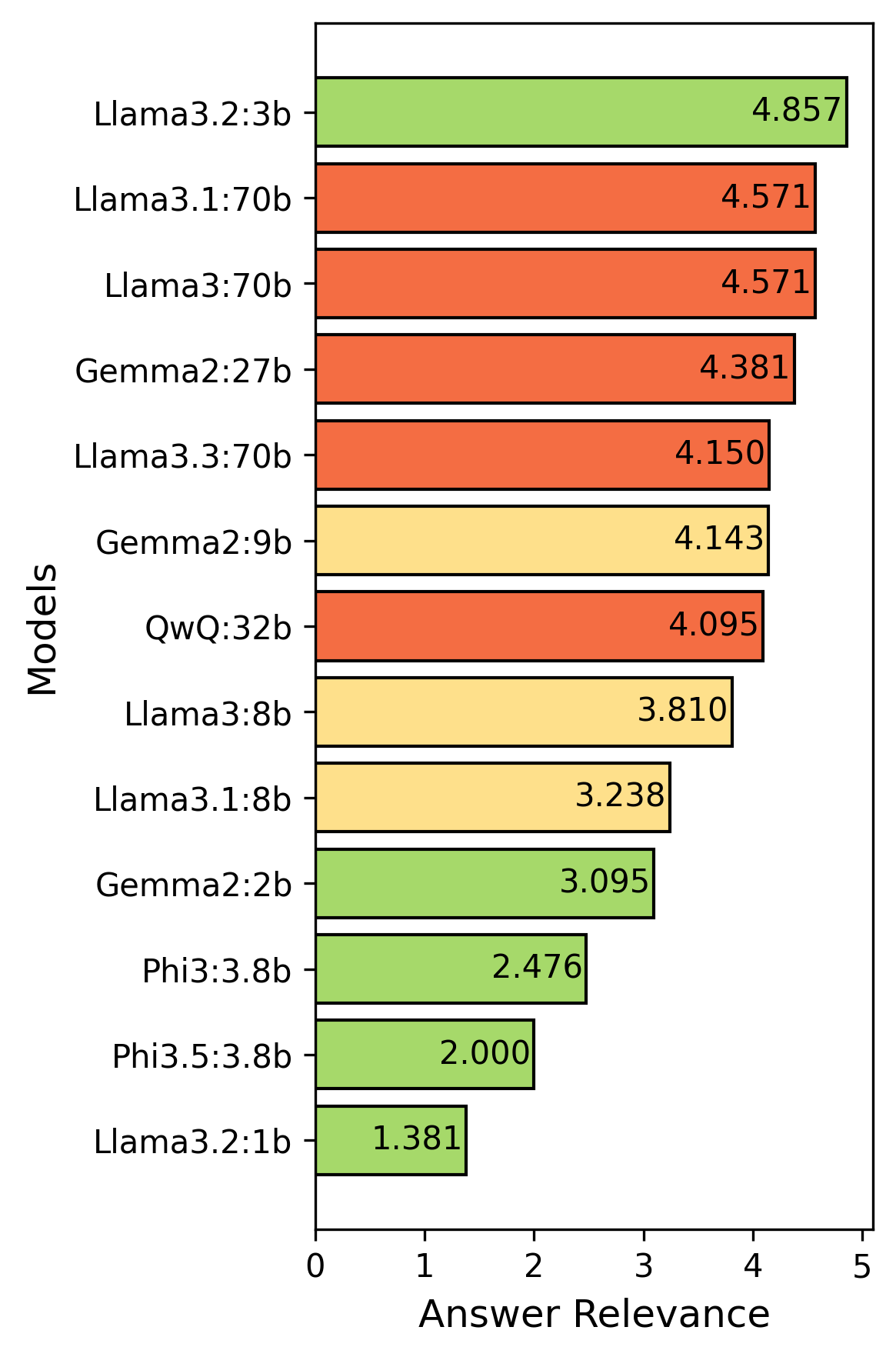} 
\caption{Answer Relevance by the 13 LLMs.}
\label{fig.3}
\end{figure}

\subsection{Experimental Results}
The experimental evaluation assessed the capacity of a variety of large language models (LLMs) to respond to queries related to quranic studies. The models were assessed based on three critical metrics: context relevance, answer faithfulness, and answer relevance. The models were categorized into three categories based on their parameter sizes: large models (marked in red), medium models (marked in yellow), and small models (marked in green). Therefore, the results were analyzed. The following is a comprehensive analysis of the performance of each category.

\subsubsection{Context Relevance}
Context relevance as shown in Figure~\ref{fig.1} evaluates how well the generated responses align with the query's context.
\begin{itemize}
    \item \textbf{Large Models (Red)}: The large models outperformed other categories, with Llama3.3:70b achieving the highest score of 0.583, followed by Llama3.2:3b (0.508), despite being categorized as a small model. Both Llama3.1:70b and Gemma2:27b achieved competitive scores of 0.492 and 0.429, respectively, while QwQ:32b recorded 0.397. These models excel at retrieving relevant information and aligning responses with the query intent due to their larger parameter size.
    \item \textbf{Medium Models (Yellow)}: Among the medium models, Gemma2:9b performed best with a score of 0.492, comparable to some large models. Llama3:8b and Llama3.1:8b followed with scores of 0.381 and 0.317, respectively. These models demonstrated decent performance but lagged behind the large models in handling complex or nuanced queries.
    \item \textbf{Small Models (Green)}: The small models struggled overall, with Llama3.2:1b achieving the lowest score of 0.254. Phi3.5:3.8b and Phi3:3.8b performed moderately with scores of 0.333, while Gemma2:2b achieved 0.413. Notably, Llama3.2:3b outperformed all expectations with a score of 0.508, surpassing even some medium and large models.
\end{itemize}

\subsubsection{Answer Faithfulness}
Answer faithfulness as shown in Figure~\ref{fig.2} measures whether the responses remain consistent with the retrieved content, avoiding inaccuracies or hallucinations.
\begin{itemize}
    \item \textbf{Large Models (Red)}: The large models dominated this metric, with Llama3.2:3b (exceptionally performing despite its small category) achieving the top score of 4.619. Llama3:70b and Llama3.1:70b scored 4.571 and 4.476, respectively, while Gemma2:27b and QwQ:32b followed with scores of 4.238 and 4.095. Their ability to maintain faithfulness highlights the advantages of larger parameter sizes.
    \item \textbf{Medium Models (Yellow)}: The medium models showed reliable performance, with Gemma2:9b scoring 4.143 and Llama3:8b achieving 3.762. Llama3.1:8b, while consistent, lagged slightly behind with 3.238. These models balanced faithfulness and efficiency but struggled with queries requiring deep contextual reasoning.
    \item \textbf{Small Models (Green)}: The small models faced significant challenges. Llama3.2:1b recorded the lowest faithfulness score of 1.381, and Phi3.5:3.8b achieved 2.000, indicating frequent inconsistencies. While Phi3:3.8b scored slightly higher at 2.476, Llama3.2:3b stood out with a score of 4.619, showcasing exceptional faithfulness that rivaled larger models.
\end{itemize}

\subsubsection{Answer Relevance}
Answer relevance as shown in Figure~\ref{fig.3} assesses whether the generated responses address the query's intent effectively.
\begin{itemize}
    \item \textbf{Large Models (Red)}: The Llama3.2:3b, while classified as small, achieved the highest relevance score of 4.857, followed closely by Llama3:70b and Llama3.1:70b, both scoring 4.571. Gemma2:27b and QwQ:32b continued to perform well, with scores of 4.381 and 4.095, respectively. These models consistently delivered relevant responses aligned with the intent behind complex queries.
    \item \textbf{Medium Models (Yellow)}: The medium models provided strong performance, particularly Gemma2:9b, which achieved 4.143. Llama3:8b followed with 3.810, and Llama3.1:8b recorded a slightly lower score of 3.238. These models addressed moderately complex queries effectively but occasionally lacked depth.
    \item \textbf{Small Models (Green)}: The small models exhibited varied performance, with Llama3.2:1b scoring the lowest at 1.381. Phi3.5:3.8b and Phi3:3.8b recorded scores of 2.000 and 2.476, respectively, indicating challenges in providing fully relevant responses. However, Llama3.2:3b once again stood out, achieving the highest score of 4.857, performing on par with the best large models.
\end{itemize}

\subsubsection{Intercoder Agreement}
We assessed intercoder agreement, which measures the extent to which evaluators within the same group report the same evaluation for a given instance. To compute this metric, we examined the percentage of instances where both evaluators assigned identical evaluation scores independently. This measure enabled us to evaluate the degree of concordance between evaluators and assess their consistency in evaluation.

\begin{table}[!ht]
\caption{The Kappa Values For The Evaluation}
\label{tab:5}
\centering
\begin{tabular}{l l l}
\hline
Creator & Model & Evaluator \\
\hline
     Meta & Llama 3.3 70B &       0.80 \\
     Meta &  Llama 3.2 3B &       0.90 \\
     Meta &  Llama 3.2 1B &       0.82 \\
     Meta & Llama 3.1 70B &       0.82 \\
     Meta &  Llama 3.1 8B &       0.85 \\
     Meta &   Llama 3 70B &       0.92 \\
     Meta &    Llama 3 8B &       0.80 \\
   Google &   Gemma 2 27B &       0.90 \\
   Google &    Gemma 2 9B &       0.92 \\
   Google &    Gemma 2 2B &       0.83 \\
Microsoft &  Phi 3.5 3.8B &       0.83 \\
Microsoft &  Phi 3.3 3.8B &       0.93 \\
  Alibaba &       QwQ 32B &       0.89 \\
\hline
\end{tabular}
\end{table}

In Table~\ref{tab:5}, we delve into the inter-annotator agreement analysis, which measures the level of agreement between human evaluators across different models using Fleiss’ Kappa. This indicates a similar level of agreement between the evaluators. 

\section{Discussion}
This section discusses the experimental findings, analyzing the performance of various LLMs categorized into large, medium, and small models across the three evaluation metrics: context relevance, answer faithfulness, and answer relevance \cite{gao2023retrieval}. This study examines the relationship among model size, response quality, and computational trade-offs, as well as the behavior of models within the Retrieval-Augmented Generation (RAG) framework.

\subsection{Performance Insights Based on Model Size}
The experimental results show that the quality of the responses across all three evaluation criteria is significantly affected by the model size.

\begin{itemize}
    \item \textbf{Large Models (Red)}: Large models, including Llama3:70b, Llama3.1:70b, Llama3.3:70b, Gemma2:27b, and QwQ:32b, consistently demonstrate superior performance compared to medium and small models. The models demonstrated superior context relevance due to their larger parameter sizes \cite{badshah2024quantifying}, which facilitate a deeper semantic understanding and enhance their ability to retrieve and integrate pertinent information. Furthermore, their enhanced performance in answer faithfulness and relevance indicates that large models are more adept at minimizing hallucinations \cite{tonmoy2024comprehensive} and producing responses that accurately address user queries. However, their computational demands remain a major trade-off, requiring substantial memory and processing power. This limits their accessibility in resource-constrained environments, making them more suitable for high-performance systems.
    \item \textbf{Medium Models (Yellow)}: Medium-sized models like Gemma2:9b, Llama3:8b, and Llama3.1:8b demonstrated strong performance relative to their parameter sizes, particularly in answer faithfulness and relevance. These models provide a balance between computational efficiency and response quality, making them ideal for systems where resources are limited but accuracy cannot be compromised. However, their performance in context relevance was slightly lower than that of large models, indicating limitations in capturing deeper relationships within the data. Medium models present a viable option for applications requiring moderate precision while maintaining resource efficiency.
    \item \textbf{Small Models (Green)}: The small models, including Llama3.2:3b, Llama3.2:1b, Gemma2:2b, Phi3:3.8b, and Phi3.5:3.8b, faced significant challenges in delivering high-quality responses. Models like Llama3.2:1b and Phi3.5:3.8b scored the lowest across all metrics, reflecting their inability to process complex queries effectively due to their smaller parameter size. Interestingly, Llama3.2:3b emerged as an outlier, achieving performance levels comparable to the large models, particularly in answer faithfulness (4.619) and answer relevance (4.857). This unexpected performance demonstrates the efficacy of architectural improvements and pre-training techniques, even on smaller models. Although small models are computationally efficient and well-suited for lightweight tasks \cite{chen2024role}, their overall limitations render them less suitable for complex queries that necessitate a deep comprehension of semantics.
\end{itemize}

\subsection{Effectiveness of the RAG Framework}
The implementation of the Retrieval-Augmented Generation (RAG) framework significantly enhanced the response quality of the evaluated models \cite{su2024implementing}. By incorporating external knowledge from the descriptive Qur’anic dataset, the models can retrieve pertinent information prior to formulating responses. This method alleviated the prevalent issue of hallucination, where language models produce factually inaccurate or irrelevant responses.

The results demonstrate that the larger models derived the greatest advantage from the RAG framework, as their higher parameter sizes facilitated more effective integration of the retrieved context, resulting in enhanced performance across all metrics. Medium and small models showed enhancements, nevertheless, their capacity limitation affected the integration of retrieved knowledge, leading to lower context alignment and fewer accurate responses.

\subsection{Trade-Off Between Computational Resources and Response Quality}
The results of this research highlight the trade-off between response quality and computational efficiency throughout several model sizes.
\begin{itemize}
    \item \textbf{Large Models} deliver outstanding performance but they are less practical for real-time or cost-sensitive installations since they need large computational resources.
    \item \textbf{Medium Models} fit for uses with intermediate hardware availability since they offer a useful compromise between dependability of performance and low resource usage.
    \item \textbf{Small Models} are quick and light-weight, yet they usually underperform on jobs needing sophisticated thinking. Nonetheless, the unexpected findings in Llama3.2:3b suggest that smaller models can still show good performance.
\end{itemize}

This trade-off emphasizes the need of choosing the suitable model size depending on particular application criteria including accuracy, speed, and resource availability.

\subsection{Surprising Performance of Llama3.2:3b}
This study's outstanding discovery is the exceptional performance of Llama3.2:3b, despite its classification as a small model. It outperformed a number of medium and even big models, achieving the highest results in answer faithfulness and answer relevance. According to this finding, factors including pre-training quality, data efficiency, and architectural upgrades can have an impact on model performance, in addition to parameter size. The potential for smaller models to produce high-quality outputs in resource-efficient environments is underscored by the robust performance of Llama3.2:3b when used in conjunction with effective frameworks such as RAG.

\subsection{Implications for Domain-Specific Tasks}
The research emphasizes both the difficulties and potential benefits of utilizing general-purpose LLMs for specialized tasks, including responding to inquiries about quranic studies. Although large models excelled in aligning responses with the given dataset, their effectiveness is significantly dependent on the quality and organization of the retrieved content. This research demonstrates how important it is to add domain-specific knowledge \cite{zhou2024gastrobot}, like curated descriptive datasets, to improve the models' abilities and lower the risk of hallucinations. The RAG framework proved to be an effective technique for ensuring that responses were contextually correct and based on credible sources.

The experimental results offer important insights into the performance of LLMs of different sizes. Large models provide exceptional accuracy and relevance, though they require significant resources, whereas medium models offer a practical compromise between performance and efficiency. Smaller models, while typically less powerful, demonstrated surprising potential, especially Llama3.2:3b, which excelled across various metrics. The use of the RAG framework enhanced the models' performance by reducing hallucinations and grounding responses in reliable data. These findings highlight the importance of model selection, optimization strategies, and retrieval mechanisms when applying LLMs to domain-specific tasks. Future work will include conducting an ablation study to compare the performance of models with and without the RAG framework, evaluating additional large language models (LLMs), and leveraging LLMs for automatic evaluation of answer faithfulness and answer relevance metrics. Additional fine-tuning strategies and assessments in other specialized domains can also be explored.

\section{Conclusions}
This study evaluated multiple large language models (LLMs) of different sizes in responding to Quranic studies-related queries using a Retrieval-Augmented Generation (RAG) framework. The findings indicate that large models, such as Llama3:70b, Llama3.1:70b, and Gemma2:27b, consistently delivered superior performance in context relevance, answer faithfulness, and answer relevance. However, their computational demands pose challenges for practical deployment. Medium-sized models, including Gemma2:9b and Llama3:8b, demonstrated a balance between efficiency and performance, making them suitable for moderately complex tasks. Interestingly, Llama3.2:3b, a small model, performed comparably to larger models in certain aspects, particularly in answer faithfulness and relevance, suggesting that architectural optimizations can enhance the capabilities of smaller models.

The study also highlights the importance of the RAG framework in improving response quality by grounding answers in external domain-specific knowledge, reducing hallucinations, and ensuring more reliable outputs. These findings emphasize the trade-offs between model size, performance, and computational efficiency, indicating that while large models are ideal for high-accuracy tasks, smaller models, when optimized, can serve as viable alternatives. Future research can focus on further optimizations, fine-tuning strategies, and expanding dataset diversity to enhance model performance across different applications.

\section*{Acknowledgment}
This research is supported by Universitas Islam Riau.



\bibliographystyle{IEEEtran}
\bibliography{reference}

\end{document}